\documentclass[letterpaper,journal]{IEEEtran}
\usepackage{amsmath,amsfonts}
\usepackage{algorithmic}
\usepackage{algorithm}
\usepackage{array}
\usepackage[caption=false,font=normalsize,labelfont=sf,textfont=sf]{subfig}
\usepackage{textcomp}
\usepackage{stfloats}
\usepackage{url}
\usepackage{verbatim}
\usepackage{graphicx}
\usepackage[nocompress]{cite}
\usepackage{multirow}
\usepackage{xcolor}
\usepackage{amssymb} 
\usepackage{booktabs} 
\usepackage{makecell}

\usepackage[hidelinks]{hyperref} 
\hypersetup{
            colorlinks=true,
            linkcolor=blue,
            anchorcolor=blue,
            citecolor=blue
            }

\begin{document}

\title{WILD-SAM: Phase-Aware Expert Adaptation of SAM for Landslide Detection in Wrapped InSAR Interferograms}

\author{Yucheng Pan, Heping Li,  Zhangle Liu, Sajid Hussain, and Bin Pan%
\thanks{Yucheng Pan, Heping Li,  Sajid Hussain and Zhangle Liu are with the School of Remote Sensing and Information Engineering, Wuhan University, Wuhan 430079, China (e-mail:  yuchengpan@whu.edu.cn; hepingli@whu.edu.cn; hussain.sajid@whu.edu.cn; liuzhangle@whu.edu.cn).}%
\thanks{Bin Pan is with the School of Remote Sensing and Information Engineering, Wuhan University, Wuhan 430079, China, and also with the Hubei Luojia Laboratory, Wuhan 430079, China (e-mail: panbin@whu.edu.cn).}%
}


\maketitle

\begin{abstract}
Detecting slow-moving landslides directly from wrapped Interferometric Synthetic Aperture Radar (InSAR) interferograms is crucial for efficient geohazard monitoring, yet it remains fundamentally challenged by severe phase ambiguity and complex coherence noise. While the Segment Anything Model (SAM) offers a powerful foundation for segmentation, its direct transfer to wrapped phase data is hindered by a profound spectral domain shift, which suppresses the high-frequency fringes essential for boundary delineation. To bridge this gap, we propose WILD-SAM, a novel parameter-efficient fine-tuning framework specifically designed to adapt SAM for high-precision landslide detection on wrapped interferograms. Specifically, the architecture integrates a Phase-Aware Mixture-of-Experts (PA-MoE) Adapter into the frozen encoder to align spectral distributions and introduces a Wavelet-Guided Subband Enhancement (WGSE) strategy to generate frequency-aware dense prompts. The PA-MoE Adapter exploits a dynamic routing mechanism across heterogeneous convolutional experts to adaptively aggregate multi-scale spectral-textural priors, effectively aligning the distribution discrepancy between natural images and interferometric phase data. Meanwhile, the WGSE strategy leverages discrete wavelet transforms to explicitly disentangle high-frequency subbands and refine directional phase textures, injecting these structural cues as dense prompts to ensure topological integrity along sharp landslide boundaries. Extensive experiments on the ISSLIDE and ISSLIDE+ benchmarks demonstrate that WILD-SAM achieves state-of-the-art performance, significantly outperforming existing methods in both target completeness and contour fidelity.
\end{abstract}


\section{Introduction}
\IEEEPARstart{S}{low-moving}  landslides constitute a critical global geological hazard, where complex geological and hydrological interactions drive continuous, subtle deformations that cause insidious yet substantial damage to infrastructure and communities over time. While traditional optical remote sensing struggles to identify subtle slope deformations, Multitemporal Interferometric Synthetic Aperture Radar (MT-InSAR) has proven to be an ideal technique for monitoring these phenomena with millimeter-level precision \cite{1221835,1315820,5765671}. Consequently, various innovative time-series InSAR strategies have been developed recently to enhance the robustness and coverage of wide-area landslide inventories \cite{WANG2023103224,11077362,rs17081462}. However, processing massive interferometric datasets using these traditional approaches is inherently labor-intensive, computationally prohibitive, and severely constrained by an over-reliance on expert interpretation. Therefore, bypassing complex conventional processing pipelines to directly and accurately extract high-frequency phase features from wrapped interferograms using advanced deep learning models is the key to achieving efficient, large-scale, and high-precision landslide detection.

Driven by the demand for efficient and intelligent monitoring, deep learning has been widely applied to InSAR-based slow-moving landslide detection. Currently, these approaches generally fall into two main categories: object detection and semantic segmentation. Object detection-based methods, highlighted by recent advances such as MSIDNet \cite{10699390} alongside other architectures \cite{rs14112690, CAI2023103516}, typically employ InSAR-derived phase gradient maps or velocity maps to automatically localize potential landslides. However, while these detection models are efficient for rapid inventory updates, they yield coarse bounding boxes that lack precise geometric details. Consequently, various semantic segmentation networks \cite{rs14081848, CAI2025104882, CAI2025104387} have been proposed to accurately delineate landslide boundaries and capture morphological characteristics. Recently, vision foundation models like DP-SAM \cite{HE2025104407} have adapted the Segment Anything Model (SAM) \cite{Kirillov2023SegmentA} for displacement segmentation. Although these architectural advancements have significantly advanced automated landslide inventorying, they are fundamentally constrained by their reliance on unwrapped deformation products, which introduces phase-unwrapping errors and prohibitively high computational costs.
\begin{figure}[t]
    \centering
    \includegraphics[width=1\linewidth]{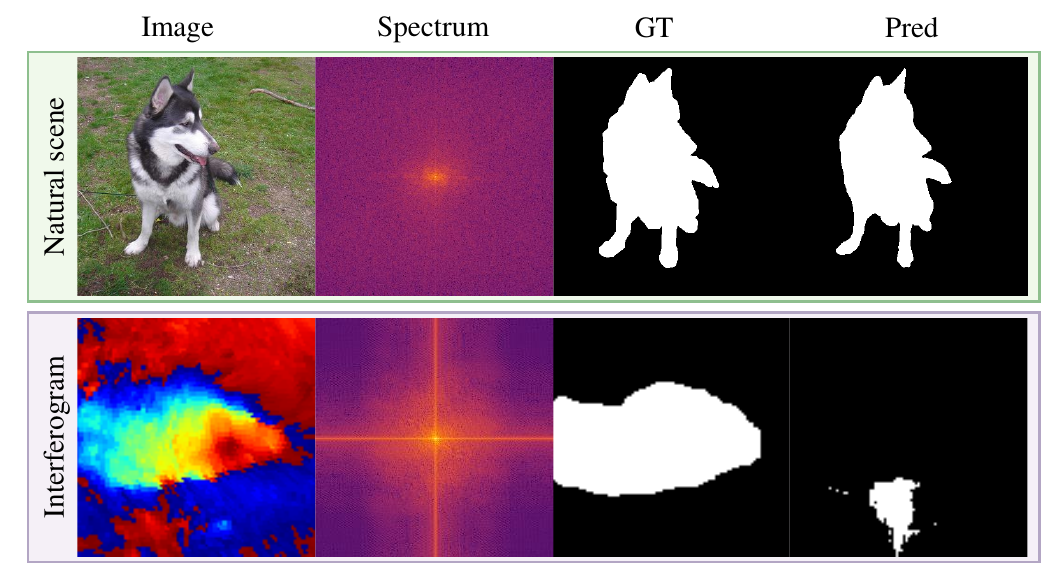}
\caption{Illustration of the domain gap between interferometric and natural data. The left column compares the visual appearance of a wrapped interferogram with a natural scene. The middle column contrasts their spectral distributions, highlighting the high-frequency nature of phase fringes. The right columns show the Ground Truth (GT) and the prediction (Pred) of the vanilla SAM with a frozen encoder, emphasizing the challenge of direct transfer.}
%\vspace{-.2cm}
\label{fig:1}
\end{figure}

To circumvent these data-level bottlenecks, recent research has shifted toward learning directly from more accessible wrapped interferograms \cite{10433141}. Specialized architectures like MB-Net \cite{ZHANG2024104300} and the explainability-constrained ECSPLAIN framework \cite{11108247} have demonstrated the potential of this paradigm. Nevertheless, these small-scale networks inherently lack the representational capacity and boundary fidelity offered by large-scale foundation models. While transferring the robust segmentation capabilities of SAM to wrapped interferograms is highly appealing, a direct deployment yields severely suboptimal results due to a profound spectral domain shift. Unlike natural images that concentrate energy in low-frequency bands, interferograms encode physical deformations in dense, high-frequency phase fringes submerged in variable coherence noise. Consequently, the inherent "low-frequency bias" of Vision Transformers (ViTs) tends to smooth out these critical phase discontinuities, leading the vanilla SAM to produce erroneous predictions with severe topological fragmentation.

To address the challenges of profound spectral domain shifts and the low-pass filtering effect of ViTs, we propose WILD-SAM, a novel parameter-efficient fine-tuning framework specifically adapted for wrapped interferometric landslide detection. Instead of fully retraining the foundation model, our architecture freezes the SAM backbone and explicitly models the physical characteristics of phase data through two tailored lightweight modules. Specifically, we integrate a Phase-Aware Mixture-of-Experts (PA-MoE) Adapter into the encoder to bridge the domain gap. PA-MoE utilizes a dynamic routing mechanism among heterogeneous convolutional experts to adaptively aggregate multi-scale spectral-textural priors, ensuring the model accurately perceives complex deformation signals. Simultaneously, we design a Wavelet-Guided Subband Enhancement (WGSE) strategy that leverages discrete wavelet transforms to disentangle high-frequency subbands. By injecting these refined features as semantic-aware dense prompts, WGSE explicitly guides the mask decoder to recover suppressed phase discontinuities and ensure topological integrity along sharp boundaries. Extensive evaluations on the ISSLIDE and ISSLIDE+ benchmarks demonstrate that WILD-SAM achieves state-of-the-art performance, significantly outperforming existing methods in both target completeness and contour fidelity.

Our main contributions are summarized as follows:
\begin{enumerate}
    \item We propose WILD-SAM, a novel framework that extends the capabilities of SAM to the challenging task of interferometric landslide detection, effectively bridging the spectral domain gap between natural images and phase data.
    \item We introduce the PA-MoE Adapter, a lightweight module that utilizes dynamic routing among heterogeneous experts to capture complex spectral characteristics and multi-scale deformation patterns specific to InSAR data.
    \item We design the WGSE strategy as a dense prompt generator, which leverages wavelet decomposition to recover fine-grained boundary details suppressed by the transformer backbone, thereby ensuring topological integrity.
    \item Extensive experiments on large-scale benchmarks demonstrate that WILD-SAM significantly outperforms SOTA methods, while maintaining robust cross-region generalization on unseen test sites.
\end{enumerate}

The remainder of this paper is organized as follows. Section \ref{sec:rw} reviews the related work on the adaptation of vision foundation models, Mixture-of-Experts in computer vision, and frequency-domain learning. Section \ref{sec:method} details the proposed WILD-SAM framework, including the technical designs of the PA-MoE Adapter and the WGSE strategy. Section \ref{sec:exp} presents extensive experimental results on various benchmarks, along with comprehensive ablation studies and visualization analysis. Finally, Section \ref{sec:con} concludes this work and discusses potential future research directions.
\section{Related Work}\label{sec:rw}

\subsection{Adaptation of Vision Foundation Models in Remote Sensing}

SAM demonstrates strong zero-shot generalization in natural images \cite{WU2025103547,SAMRS,10944021}. Similar generalization trends are also observed in controllable generation and editing models \cite{shen2025imagedit, shen2025imagharmony, shen2024imagpose}. However, direct transfer to remote sensing fails due to severe domain gaps \cite{OSCO2023103540}. This mismatch is especially pronounced when the input departs from natural-image statistics and contains task-specific physical patterns. As a result, directly reusing off-the-shelf foundation models often leads to unstable representations and degraded segmentation quality.
PEFT becomes the dominant solution \cite{pmlr-v97-houlsby19a, hu2022lora}. LoRA-based methods \cite{10637992,10551868} provide lightweight adaptation but struggle with complex radar patterns. Adapter-based methods \cite{chen2023sam,10.1007/978-981-96-0966-6_17} offer stronger flexibility, with extensions such as SCD-SAM \cite{10543161} and UAdapter \cite{WEI2024446}. Further task-specific designs include ClassWise-SAM-Adapter \cite{10849617} and SAMSR \cite{TIAN2025114775}. These methods confirm that lightweight adaptation is effective when the target domain can be aligned by modest architectural changes. However, they are still mainly designed for optical or conventional SAR settings.
Architectural redesign further improves performance, including MeSAM \cite{10522788}, BSDSNet \cite{rs16071150}, and SAM-CFFNet \cite{rs16132334}. These methods improve multi-scale modeling or task-specific decoding, but all focus on optical or amplitude SAR data. Adaptation to interferometric phase data remains unexplored. Wrapped interferograms exhibit fundamentally different spectral and geometric properties, making their transfer challenge more severe than standard remote sensing adaptation. Therefore, we propose WILD-SAM to bridge this gap.
\begin{figure*}[t!]
    \centering
    \includegraphics[width=\textwidth]{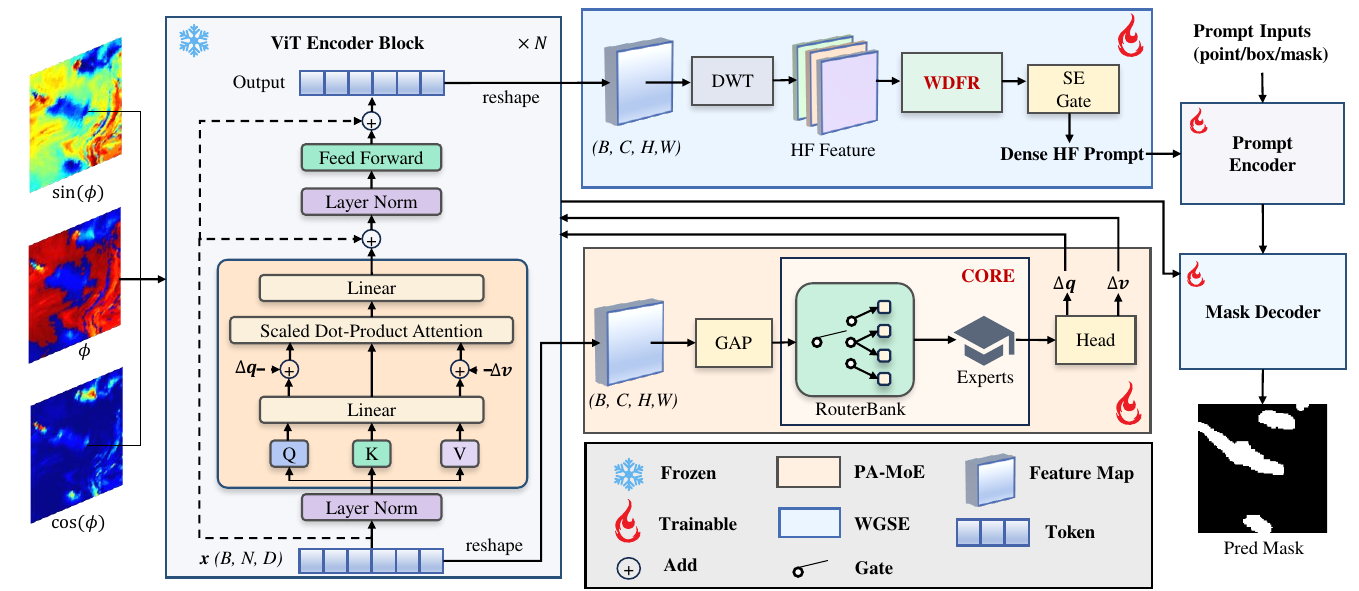}
    \caption{Overview of the proposed WILD-SAM architecture. The framework takes a three-channel phase representation as input and consists of a frozen ViT encoder block (left) alongside two tailored lightweight modules: the Phase-Aware Mixture-of-Experts (PA-MoE) Adapter (bottom), driven by the Convolutional Routing Experts (CORE) for adaptive feature modulation, and the Wavelet-Guided Subband Enhancement (WGSE) strategy (top), which utilizes a Wavelet-Domain Feature Rectifier (WDFR) to generate dense high-frequency prompts. The Mask Decoder (right) predicts precise landslide segmentation guided by these prompts.}
    \label{fig:2}
\end{figure*}
\subsection{Mixture of Experts in Computer Vision}

MoE enables conditional computation for heterogeneous feature modeling \cite{6797059,Shazeer2017OutrageouslyLN}. Efficient routing strategies include Switch Transformers \cite{fedus2022switch}, Expert Choice \cite{NEURIPS2022_2f00ecd7}, and Tutel \cite{hwang2023tutel}. Extensions to vision include V-MoE \cite{riquelme2021scaling} and patch-level routing \cite{chowdhury2023patch}, while Soft MoE \cite{puigcerver2024from} improves stability. These studies show that expert routing is particularly effective when the input contains diverse local patterns that cannot be handled uniformly. This property is highly relevant to interferometric data, where deformation signals, fringes, and noise often coexist within the same scene.
Recent advances further enhance expert selection and robustness \cite{dai-etal-2024-deepseekmoe,komatsuzaki2023sparse,dou-etal-2024-loramoe,hazimeh2021dselect,zhang2023robust}. Similar conditional modeling ideas also appear in controllable generation \cite{shen2025boosting,shen2025imagedit}. These methods improve specialization, routing flexibility, and robustness under complex feature distributions. However, existing MoE designs do not consider phase-specific characteristics. They are not designed to distinguish deformation-related structures from coherence noise or to model the anisotropic patterns of interferometric fringes. Therefore, we design PA-MoE for phase-aware feature routing.

\subsection{Frequency Domain Learning for Deep Neural Networks}

ViTs exhibit a strong low-frequency bias \cite{10163861,10378515}. This suppresses high-frequency details critical for interferometric phase analysis \cite{10163861}. Recovering high-frequency information is therefore essential. This issue is particularly critical in wrapped interferograms, where fine phase discontinuities directly determine boundary quality. Without explicit compensation, transformer features tend to over-smooth these structures and produce fragmented predictions.
Frequency-domain methods address this limitation via explicit decomposition. Wavelet-based transformers \cite{10.1007/978-3-031-19806-9_19,10657988} preserve multi-scale structures. Frequency-aware fusion methods \cite{chen2024frequency} enhance boundary details. SAR-specific designs \cite{s24237821,electronics13030490} further validate frequency decomposition. These studies consistently show that separating low- and high-frequency components improves structural fidelity and noise robustness. They also suggest that frequency priors are especially useful when fine boundaries are easily overwhelmed by dominant low-frequency semantics.
Similar structure-preserving modeling also appears in controllable generation tasks \cite{shen2025imaggarment,shen2025imagdressing,shen2025longterm}. However, these ideas are rarely integrated with foundation models for phase data. More importantly, existing methods seldom convert recovered frequency cues into explicit guidance for downstream decoding. Therefore, we propose WGSE to explicitly enhance high-frequency phase representations.

\section{Method}\label{sec:method}
In this section, the proposed WILD-SAM is elaborated. Specifically, the overall pipeline and components of the proposed framework are briefly illustrated in Fig. \ref{fig:2}. Then, the proposed components are introduced for wrapped interferometric landslide detection. Finally, the loss functions are constructed in detail.
\subsection{Overview}

The adaptation of SAM to wrapped interferograms is challenged by two fundamental issues. First, a severe spectral domain gap exists between natural images and interferometric phase data, causing the frozen encoder to inadequately perceive phase-specific patterns. Second, the inherent low-frequency bias of ViTs tends to suppress high-frequency phase discontinuities, which are critical for precise landslide boundary delineation. To address these two issues in a unified and parameter-efficient manner, we propose WILD-SAM, which augments a frozen SAM backbone with two lightweight yet complementary modules, namely the Phase-Aware Mixture-of-Experts (PA-MoE) Adapter and the Wavelet-Guided Subband Enhancement (WGSE) strategy.

As illustrated in Fig.~\ref{fig:2}, WILD-SAM takes a wrapped interferogram $\phi$ as input and first constructs a three-channel representation,
\begin{equation}
\mathbf{I} = [\phi, \sin(\phi), \cos(\phi)],
\end{equation}
where $\sin(\phi)$ and $\cos(\phi)$ are introduced to alleviate the discontinuity caused by phase wrapping. The resulting representation $\mathbf{I}$ is then fed into the Patch Embedding layer and sequentially processed by the frozen ViT encoder of SAM.

To reduce the spectral mismatch between natural-image pretraining and phase-data perception, we insert the PA-MoE Adapter in parallel with the multi-head attention blocks. This module extracts intermediate spatial features, dynamically routes them through heterogeneous experts, and injects adaptive perturbations into the attention projections. In this way, the frozen backbone can selectively adapt to phase-specific statistics without requiring full fine-tuning.

Meanwhile, to compensate for the loss of high-frequency structures induced by the low-pass behavior of the ViT encoder, WGSE operates between the encoder and the prompt space. It decomposes intermediate features into frequency subbands, enhances the high-frequency components, and transforms them into a dense prompt. This prompt is then fed into the SAM Prompt Encoder together with optional sparse prompts, enabling the Mask Decoder to recover sharper boundaries and more coherent landslide topologies.

By jointly addressing feature misalignment in the encoder and structural detail loss in the decoder guidance, WILD-SAM achieves accurate and topologically consistent segmentation on wrapped interferograms.

\subsection{Phase-Aware Mixture-of-Experts}

Although SAM exhibits remarkable transferability on natural images, its direct adaptation to wrapped interferograms remains highly nontrivial. Unlike natural scenes, phase data contain complex coherence noise, anisotropic fringes, and heterogeneous deformation patterns, all of which differ substantially from the statistics learned during natural-image pretraining. Directly reusing the frozen encoder therefore yields insufficient phase awareness, whereas full fine-tuning can easily damage the pretrained generalization capability. Motivated by this observation, we introduce the PA-MoE Adapter, which performs lightweight and conditional feature modulation to bridge the spectral domain gap while preserving the backbone parameters.

Given an intermediate spatial feature map $\mathbf{X} \in \mathbb{R}^{B \times C \times H \times W}$, where $B$, $C$, $H$, and $W$ denote the batch size, channel number, height, and width, respectively, PA-MoE extracts phase-aware priors through the Convolutional Routing Experts (CORE) module. Specifically, CORE consists of four heterogeneous experts that capture complementary phase characteristics. The first expert $E_1(\cdot)$ focuses on local textural patterns, the second expert $E_2(\cdot)$ enlarges the receptive field to encode broader deformation context, the third expert $E_3(\cdot)$ strengthens directional fringe structures, and the fourth expert $E_4(\cdot)$ emphasizes high-frequency discontinuities that are critical for precise boundary localization. This heterogeneous design allows the model to simultaneously represent subtle deformation cues and suppress interference from coherence noise.

To adaptively combine these experts according to the current input, we first apply Global Average Pooling to obtain a compact global descriptor
\begin{equation}
\mathbf{g} = \mathrm{GAP}(\mathbf{X}),
\end{equation}
where $\mathbf{g} \in \mathbb{R}^{C}$ summarizes the global spectral-spatial context of the input feature. Based on this descriptor, a bottleneck routing network predicts the expert weights
\begin{equation}
\mathbf{w} = \mathrm{Softmax}\left(\mathbf{W}_2 \cdot \delta(\mathbf{W}_1 \cdot \mathbf{g})\right),
\end{equation}
where $\mathbf{w} \in \mathbb{R}^{4}$ denotes the routing distribution over the four experts, $\delta(\cdot)$ is the GELU activation function, and $\mathbf{W}_1$ and $\mathbf{W}_2$ are learnable projection matrices. In this way, the router can dynamically emphasize the most relevant experts for a given interferometric pattern.

The outputs of all experts are then aggregated by soft routing as
\begin{equation}
\mathbf{Y}_{\mathrm{CORE}} = \sum_{i=1}^{4} w_i \cdot E_i(\mathbf{X}),
\end{equation}
where $w_i$ is the $i$-th element of $\mathbf{w}$ and $Y_{\mathrm{CORE}}$ denotes the fused phase-aware feature. This aggregation enables the model to adaptively balance local texture, contextual structure, directional response, and boundary-sensitive high-frequency information.
\begin{figure}[t]
    \centering
    \includegraphics[width=\columnwidth]{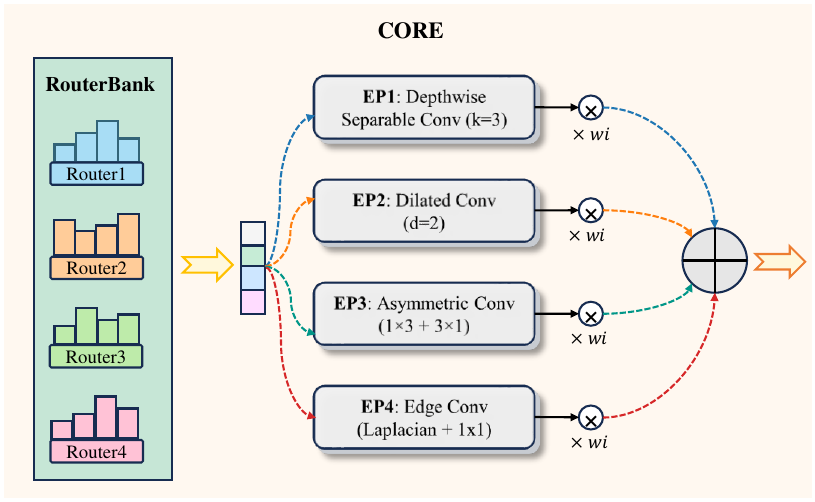}
    \caption{Details of our Convolutional Routing Experts (CORE) module. It leverages a dynamic RouterBank to aggregate multi-scale features from four heterogeneous convolutional experts: $E1$ (Depthwise), $E2$ (Dilated), $E3$ (Asymmetric), and $E4$ (Laplacian). This design adaptively distinguishes complex deformation signals from variable coherence noise.}
    \label{fig:3}
\end{figure}
After fusion, $Y_{\mathrm{CORE}}$ is projected by a $1 \times 1$ convolutional head and reshaped into token space to produce two adaptive perturbations, denoted by $\Delta q$ and $\Delta v$. These perturbations are injected into the Query and Value projections of the frozen attention block:
\begin{equation}
\mathbf{Q}' = \mathbf{Q} + \alpha \Delta \mathbf{q}, \qquad \mathbf{V}' = \mathbf{V} + \alpha \Delta \mathbf{v},
\end{equation}
where $\mathbf{Q}$ and $\mathbf{V}$ are the original Query and Value features, $\mathbf{Q}'$ and $\mathbf{V}$ are the modulated features, and $\alpha$ is a learnable scaling parameter that controls the perturbation strength. Through this residual modulation mechanism, PA-MoE enables the frozen SAM encoder to selectively adapt to phase-specific statistics without modifying the entire backbone. As a result, the encoder becomes more sensitive to interferometric textures and more robust to variable coherence noise.

\subsection{Wavelet-Guided Subband Enhancement}

Although PA-MoE improves the phase awareness of encoder features, it does not explicitly address another critical limitation of ViTs, namely their tendency to favor low-frequency structures while suppressing fine high-frequency details. In wrapped interferograms, however, landslide boundaries are often encoded by dense phase fringes and sharp discontinuities. Once these high-frequency cues are smoothed out, the decoder tends to produce blurred, incomplete, or topologically fragmented predictions. To explicitly recover such lost structures, we design the Wavelet-Guided Subband Enhancement (WGSE) strategy, which transforms high-frequency encoder features into a dense prompt for boundary-aware decoding.

Given the intermediate feature map $\mathbf{X} \in \mathbb{R}^{B \times C \times H \times W}$, WGSE first applies a Discrete Wavelet Transform (DWT) to decompose it into one low-frequency subband and three high-frequency subbands:
\begin{equation}
\mathbf{X}_{\mathrm{LL}}, \mathbf{X}_{\mathrm{LH}}, \mathbf{X}_{\mathrm{HL}}, \mathbf{X}_{\mathrm{HH}} = \mathrm{DWT}(\mathbf{X}),
\end{equation}
where $\mathbf{X}_{\mathrm{LL}}$ denotes the low-frequency approximation component, while $\mathbf{X}_{\mathrm{LH}}$, $\mathbf{X}_{\mathrm{HL}}$, and $\mathbf{X}_{\mathrm{HH}}$ correspond to the vertical, horizontal, and diagonal high-frequency subbands, respectively. Since our goal is to recover fine structural boundaries rather than reinforce already dominant low-frequency semantics, we discard $X_{LL}$ and only retain the high-frequency components for subsequent enhancement.

The retained subbands are processed by the Wavelet-Domain Feature Rectifier (WDFR), which consists of two stages. The first stage is the Anisotropic Feature Modulator (AFM), designed to capture direction-specific phase structures. Because interferometric fringes often exhibit strong anisotropy, each subband should be modeled independently instead of being prematurely fused. For each subband $\mathbf{X}_c$, where $c \in \{LH, HL, HH\}$, AFM produces a modulated feature
\begin{equation}
\mathbf{F}_c = \mathrm{AFM}(\mathbf{X}_c),
\end{equation}
where $F_c \in \mathbb{R}^{N \times D}$ denotes the tokenized representation after directional enhancement, with $N$ being the token number and $D$ the embedding dimension. This operation preserves subband-specific directional cues that are crucial for capturing fringe geometry.
\begin{figure}[t]
    \centering
    \includegraphics[width=\columnwidth]{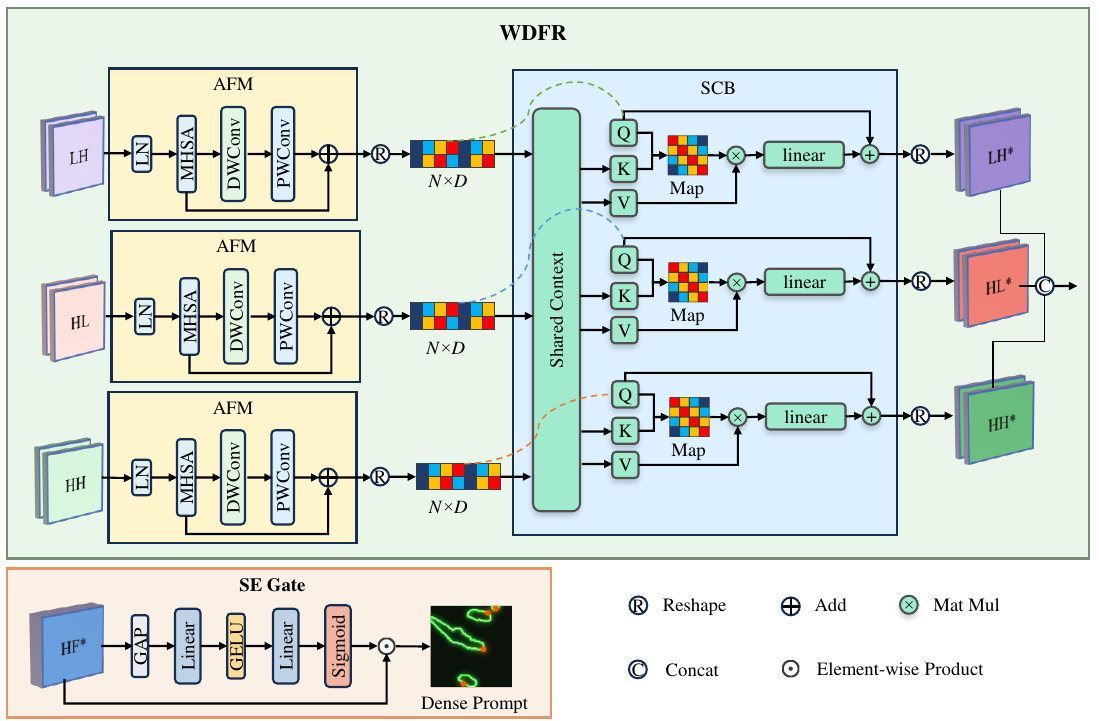}
    \caption{Wavelet-Domain Feature Rectifier (WDFR) and SE Gate. The high-frequency subbands extracted by DWT are first processed independently by Anisotropic Feature Modulators (AFM) and then structurally aligned via a Spectral Coupling Bridge (SCB) to obtain the final dense high-frequency prompt.}
    \label{fig:4}
\end{figure}
The second stage is the Spectral Coupling Bridge (SCB), which models interactions across different high-frequency subbands. Although each subband emphasizes a particular directional component, complete landslide boundaries usually emerge from the joint configuration of multiple subbands. To prevent fragmented predictions caused by isolated processing, we aggregate cross-subband information through a shared context $S$ and perform cross-attention:
\begin{equation}
\mathbf{F}^*_c = \mathbf{F}_c + \mathrm{Linear}\left(\mathrm{Softmax}\left(\frac{\mathbf{Q}_c \mathbf{K}_S^T}{\sqrt{d_k}}\right)\mathbf{V}_S\right),
\end{equation}
where $Q_c = F_c W_Q$ is the Query projected from subband feature $F_c$, $K_S = S W_K$ and $V_S = S W_V$ are the Key and Value projected from the shared context $S$, $W_Q$, $W_K$, and $W_V$ are learnable projection matrices, and $d_k$ is the key dimension used for scaling. The enhanced feature $F^*_c$ therefore preserves directional specificity while incorporating complementary structural cues from the other subbands.

After cross-band interaction, the enhanced subbands are concatenated as
\begin{equation}
\mathbf{F}^* = \mathrm{Concat}(\mathbf{F}^*_{\mathrm{LH}}, \mathbf{F}^*_{\mathrm{HL}}, \mathbf{F}^*_{\mathrm{HH}}).
\end{equation}
However, high-frequency components in interferograms contain not only useful boundary information but also substantial noise. To adaptively suppress noisy responses and highlight deformation-related structures, we further employ a Squeeze-and-Excitation gate:
\begin{equation}
\mathbf{s} = \sigma(\mathbf{W}_4 \cdot \delta(\mathbf{W}_3 \cdot \mathrm{GAP}(\mathbf{F}^*))),
\end{equation}
where $s$ denotes the channel-wise attention weight, $\sigma(\cdot)$ is the Sigmoid activation, $\delta(\cdot)$ is the GELU activation, and $W_3$ and $W_4$ are learnable weight matrices. The final dense high-frequency prompt is then generated as
\begin{equation}
\mathbf{P}_{\mathrm{HF}} = \mathbf{s} \odot \mathbf{F}^*,
\end{equation}
where $\odot$ denotes element-wise multiplication. This prompt serves as an explicit spectral guidance signal and is fed into the Prompt Encoder together with optional sparse prompts. By doing so, WGSE injects boundary-critical high-frequency information directly into the decoding stage, thereby improving contour sharpness and maintaining topological continuity.

\subsection{Loss Function}

To jointly optimize pixel-level classification accuracy and region-level structural quality, we adopt a hybrid loss composed of Binary Cross-Entropy and Dice loss:
\begin{equation}
L_{\mathrm{total}} = L_{\mathrm{BCE}} + \lambda L_{\mathrm{Dice}},
\end{equation}
where $L_{\mathrm{BCE}}$ supervises pixel-wise foreground-background discrimination, $L_{\mathrm{Dice}}$ encourages region overlap between prediction and ground truth, and $\lambda$ is a balancing coefficient. This formulation is particularly suitable for landslide detection, where foreground regions are often sparse and irregular. By combining the two terms, the model is encouraged to achieve both accurate localization and coherent shape reconstruction.E

\section{Experiment and Analysis}\label{sec:exp}
This section presents a comprehensive experimental evaluation of the proposed WILD-SAM. The experimental setup is first established by detailing the selected public InSAR datasets alongside essential implementation configurations and evaluation metrics. Subsequent comparative assessments and cross-region generalization experiments are presented to validate the overall superiority and robustness of the model against existing SOTA methods. The section concludes with extensive ablation studies to rigorously investigate the individual contributions of the proposed components.
\subsection{Datasets}
To rigorously assess the performance and cross-region robustness of WILD-SAM, we conduct evaluations on three representative InSAR landslide datasets: ISSLIDE \cite{10433141}, ISSLIDE+ \cite{11108247}, and Hunza-InSAR \cite{HUSSAIN2025103956}. These benchmarks encompass a diverse range of geological settings, phase fringe patterns, and coherence levels, providing a comprehensive platform for validating the effectiveness of our proposed framework on wrapped interferograms.

\subsubsection{ISSLIDE} The dataset is a Sentinel-1 InSAR benchmark for detecting slowly deforming areas  with expert pixel-level annotations. It provides interferograms built with 6/12/18-day temporal baselines together with coherence  and wrapped phase difference  products. The release includes 200 surface-motion phenomena from 54 interferograms and 13,230 cropped 100×100 patches,including 8,701 motion-period samples, available both as original interferograms with vector labels and as ML-ready patches with binary masks.

\subsubsection{ISSLIDE+} ISSLIDE+ is an extension of ISSLIDE and is curated to be strictly disjoint from it, providing broader coverage of slow-moving areas across larger spatial extents and more diverse scenarios. The dataset is constructed by performing a sliding-window scan over interferograms to systematically retrieve candidate moving areas. This procedure yields approximately 479 newly identified moving areas. The interferograms and their corresponding annotations are then cropped into local patches, resulting in 31,554 samples in total. Each sample is a 100×100 pixel interferogram patch paired with its corresponding moving-area annotation.

\subsubsection{Hunza-InSAR} The KKH-Lower Hunza dataset is derived from a time-series InSAR monitoring study along the Lower Hunza section of the Karakoram Highway (KKH) in northern Pakistan. The study uses Sentinel-1 SAR data with a 12-day temporal revisit and processes 60 Single Look Complex (SLC) scenes acquired in Interferometric Wide (IW) mode with VV polarization. By combining historical data and expert interpretation, 36 active landslides are identified within the study area. We use this dataset as a cross-region generalization benchmark to evaluate the robustness of the proposed model under previously unseen geological settings and observation conditions.
% ====== table ======
\begin{table*}[!t]
\centering
\caption{Quantitative comparison with SOTA methods on the ISSLIDE and ISSLIDE+ datasets.Upward arrows ($\uparrow$) indicate that higher values are better, while downward arrows ($\downarrow$) indicate that lower values are better. The first, second, and third-best values are highlighted in \textcolor{red}{red}, \textcolor{blue}{blue}, and \textcolor{green}{green}, respectively.}
\label{tab:1}
\renewcommand{\arraystretch}{1.1}
\setlength{\tabcolsep}{4.5pt}
\small
\resizebox{\textwidth}{!}{%
\begin{tabular}{c | c | ccccc | ccccc}
\noalign{\hrule height 1.5pt} 
\multirow{2}{*}{\textbf{Category}} & \multicolumn{1}{c|}{\multirow{2}{*}{\textbf{Method}}} 
& \multicolumn{5}{c|}{\textbf{ISSLIDE}} 
& \multicolumn{5}{c}{\textbf{ISSLIDE+}} \\
\cline{3-12}
              & & \textbf{Prec (\%)} & \textbf{Rec (\%)}    & \textbf{IoU} ({\scriptsize $\uparrow$}) & \textbf{Dice} ({\scriptsize $\uparrow$}) & \textbf{HD} ({\scriptsize $\downarrow$}) 
              & \textbf{Prec (\%)} & \textbf{Rec (\%)}    & \textbf{IoU} ({\scriptsize $\uparrow$}) & \textbf{Dice} ({\scriptsize $\uparrow$}) & \textbf{HD} ({\scriptsize $\downarrow$}) \\
\hline 

\multirow{4}{*}{\textbf{CNN-based}}
              & Unet\cite{ronneberger2015u}          & 35.32          & 75.67          & 0.3086          & 0.4492           & 67.71  
              & 74.18          & 68.64          & 0.4993          & 0.6421           & 48.27             \\
              & ResUnet\cite{8959021}        & 49.80          & 50.75          & 0.3233          & 0.4864           & 69.74  
              & 78.97          & 64.80          & 0.5075          & 0.6480           & 50.30             \\
              & DeepLabV3\cite{chen2017rethinking}      & 62.41          & 53.26          & 0.3888          & 0.5180           & 28.98  
              & 82.12          & 77.19          & 0.6357          & 0.7638           & 31.81             \\
              & FCN\cite{long2015fully}             & 61.64          & 54.17          & 0.3857          & 0.5242           & 27.68  
              & 84.26          & 78.44          & 0.6619          & 0.7843           & 29.58             \\
\hline

\multirow{2}{*}{\textbf{Transformer-based}}
              & SegFormer\cite{xie2021segformer}      & 59.90          & \textcolor{blue}{82.46} & 0.5313          & 0.6939           & 24.79  
              & 71.12          & 63.49          & 0.5048          & 0.6709           & 45.49             \\
              & MaskFormer\cite{cheng2021per}    & 64.97          & 72.22          & 0.5198          & 0.6465           & \textcolor{green}{23.18}  
              & 63.33          & \textcolor{red}{97.18} & 0.6219          & 0.7669           & 20.91             \\
\hline

\multirow{2}{*}{\textbf{Landslide SOTA}}
              & DSCF-LDNet\cite{11271038}        & \textcolor{green}{73.17}          & 72.00          & 0.5514          & 0.6888           & 29.97  
              & \textcolor{green}{86.46}          & 85.12          & 0.7379          & 0.8389           & 20.99             \\
              & MB-Net\cite{ZHANG2024104300}    & \textcolor{blue}{82.31} & 70.12          & \textcolor{blue}{0.6006} & \textcolor{blue}{0.7298} & \textcolor{red}{11.09} 
              & 86.05          & 87.78          & \textcolor{green}{0.7754}          & \textcolor{green}{0.8564}           & \textcolor{green}{16.75}              \\
\hline

\multirow{2}{*}{\textbf{SAM-based}}
              & RSPrompter\cite{10409216}    & 72.62          & 73.65          & 0.5596          & 0.7010           & 30.70  
              & 78.63          & 73.42          & 0.6217          & 0.7395           & 29.78             \\
              & MeSAM\cite{rs16071150}        & 66.46          & \textcolor{green}{82.42}          & \textcolor{green}{0.5642}          & \textcolor{green}{0.7034}           & 33.19  
              & \textcolor{blue}{88.09} & \textcolor{green}{89.94}          & \textcolor{blue}{0.8034} & \textcolor{blue}{0.8772} & \textcolor{blue}{14.06} \\
\hline
\textbf{Proposed}      & \textbf{Ours} & \textcolor{red}{84.78} & \textcolor{red}{87.23} & \textcolor{red}{0.7510} & \textcolor{red}{0.8526}  & \textcolor{blue}{15.33} 
              & \textcolor{red}{94.55} & \textcolor{blue}{95.69} & \textcolor{red}{0.9072} & \textcolor{red}{0.9508}  & \textcolor{red}{6.636}    \\
\noalign{\hrule height 1.5pt} 
\end{tabular}%
}
\end{table*}
\subsection{Implementation Details and Evaluation Metrics.} 
We implement our method in the PyTorch framework and conduct all experiments on four NVIDIA RTX 4090 GPUs.
The model is trained for 100 epochs with a batch size of 32.
For preprocessing, all input channels are linearly mapped to [0, 255] to match the SAM-style normalization, and inputs are resized by preserving the aspect ratio such that the longest side is 256.
We use AdamW with an initial learning rate of $1\times10^{-5}$, weight decay of 0.01, and the default hyperparameters
$\boldsymbol{\beta}=(0.9,0.999)$ and $\boldsymbol{\epsilon}=10^{-8}$.
To comprehensively evaluate the effectiveness of the proposed method, we report Precision, Recall, Intersection over Union (IoU), Dice, and the Hausdorff distance (HD) for quantitative assessment.
% ====== figure ======
\begin{figure*}[!t]
    \centering
    \includegraphics[width=\textwidth]{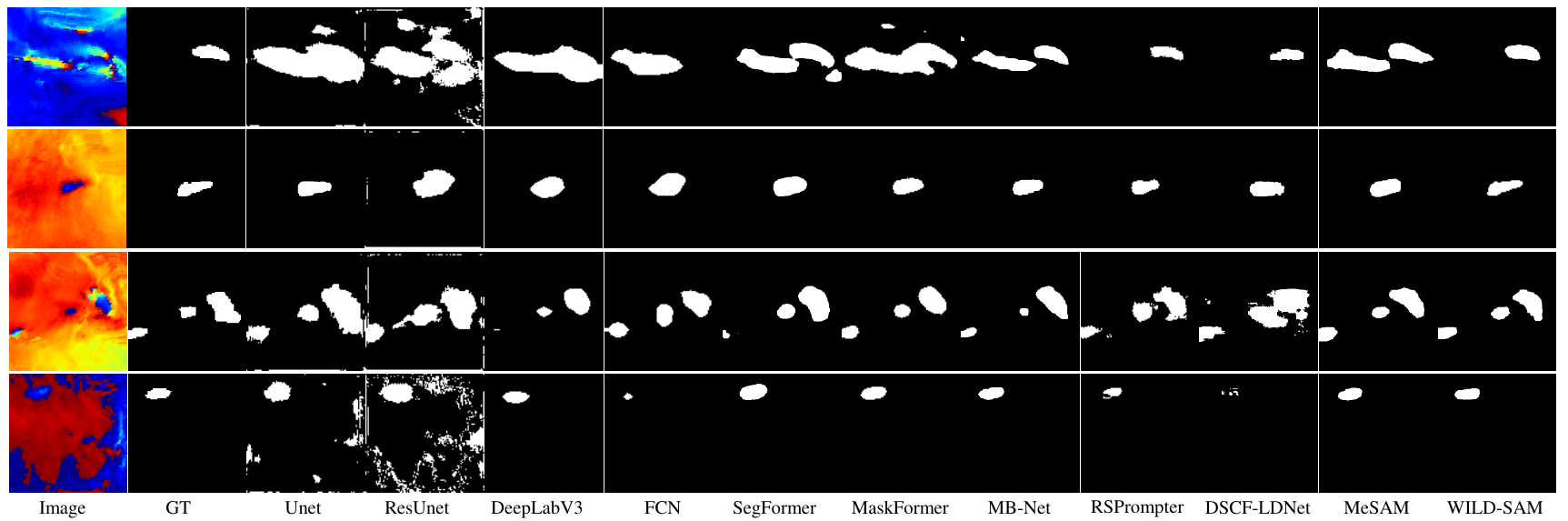}
    \caption{Qualitative comparison of segmentation results on the ISSLIDE\cite{10433141} dataset. From left to right: input wrapped interferograms, ground truth (GT), and prediction masks from various baseline models alongside the proposed WILD-SAM.}
    \label{fig:5}
\end{figure*}

\subsection{Baseline Models}
To comprehensively evaluate the performance of the proposed method, the selected baseline models are categorized into four main groups: CNN-based architectures, including UNet\cite{ronneberger2015u}, ResUNet\cite{8959021}, DeepLabv3\cite{chen2017rethinking}, and FCN\cite{long2015fully}; Transformer-based architectures, including SegFormer\cite{xie2021segformer} and MaskFormer\cite{cheng2021per}; Landslide-specific SOTA methods, including MB-Net\cite{ZHANG2024104300} and DSCF-LDNet\cite{11271038}, and SAM-based models, including RSPrompter\cite{10409216} and MeSAM\cite{rs16071150}. These diverse baselines are carefully selected to ensure a rigorous and comprehensive evaluation of the proposed framework. 
\subsection{Comparison with State-of-the-art Methods}
The quantitative comparison results of the proposed WILD-SAM and other SOTA architectures on the ISSLIDE and ISSLIDE+ datasets are summarized in Table \ref{tab:1}. As shown in the table, our proposed method consistently outperforms all competitors across the most comprehensive evaluation metrics. Among the compared baselines, traditional CNN-based architectures and Transformer-based models achieve reasonable performance, but they generally struggle with the severe spectral domain gap inherent in wrapped interferograms. These generic segmentation networks suffer from either over-smoothing bias or topological fragmentation, leading to suboptimal boundary delineation and high HD errors. 

Specifically, on the ISSLIDE dataset, WILD-SAM demonstrates superior capabilities in target localization and boundary fidelity compared to both generic baselines and specialized models. Our method achieves a Precision of 84.78\% and an IoU of 0.7510, representing a substantial improvement over the Transformer-based SegFormer, which yields a Precision of 59.90\% and an IoU of 0.5313. While the landslide-specific MB-Net demonstrates exceptional boundary precision with the lowest HD of 11.09, WILD-SAM remains highly competitive with a value of 15.33, significantly outperforming other SOTA candidates such as DSCF-LDNet and MeSAM, which result in HD errors of 29.97 and 33.19, respectively. This high geometric precision is fundamentally attributed to the WGSE strategy. By explicitly recovering high-frequency boundary details suppressed by the backbone, WGSE guides the mask decoder to focus on sharp phase discontinuities and ensures topological integrity.
\begin{figure*}[!t]
    \centering
    \includegraphics[width=\textwidth]{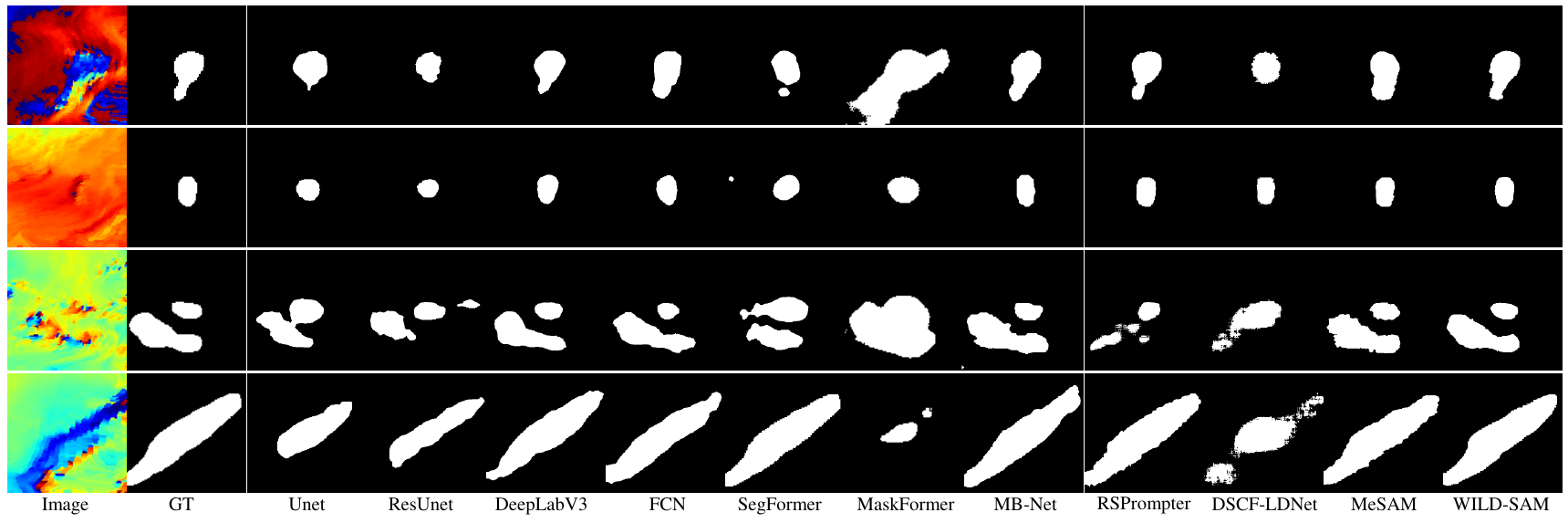}
    \caption{Qualitative comparison of segmentation results on the ISSLIDE+\cite{10433141} dataset. From left to right: input wrapped interferograms, ground truth (GT), and prediction masks from various baseline models alongside the proposed WILD-SAM.}
    \label{fig:6}
\end{figure*}

Evaluations on the larger and more diverse ISSLIDE+ dataset further confirm the robust generalization of WILD-SAM under varying geological settings. While certain competitive baselines exhibit performance imbalances, as exemplified by MaskFormer reaching a peak Recall of 97.18\% 
but a deficient Precision of 63.33\% due to excessive false positives, WILD-SAM strikes an optimal balance with an outstanding Precision of 94.55\% and a Dice score of 0.9508. Notably, our model significantly minimizes the HD to 6.636, a level of boundary fidelity that substantially outperforms specialized landslide models like MB-Net and SAM-based MeSAM, which yield HD values of 16.75 and 14.06, respectively. This stability against complex noise and variable patterns is driven by the PA-MoE Adapter. Through dynamic routing among heterogeneous experts, the PA-MoE ensures that the frozen backbone precisely perceives unique phase textures while adaptively suppressing coherence noise across diverse interferometric scenarios.
% ====== figure ======
\begin{figure*}[t]
    \centering
    \includegraphics[width=0.95\textwidth]{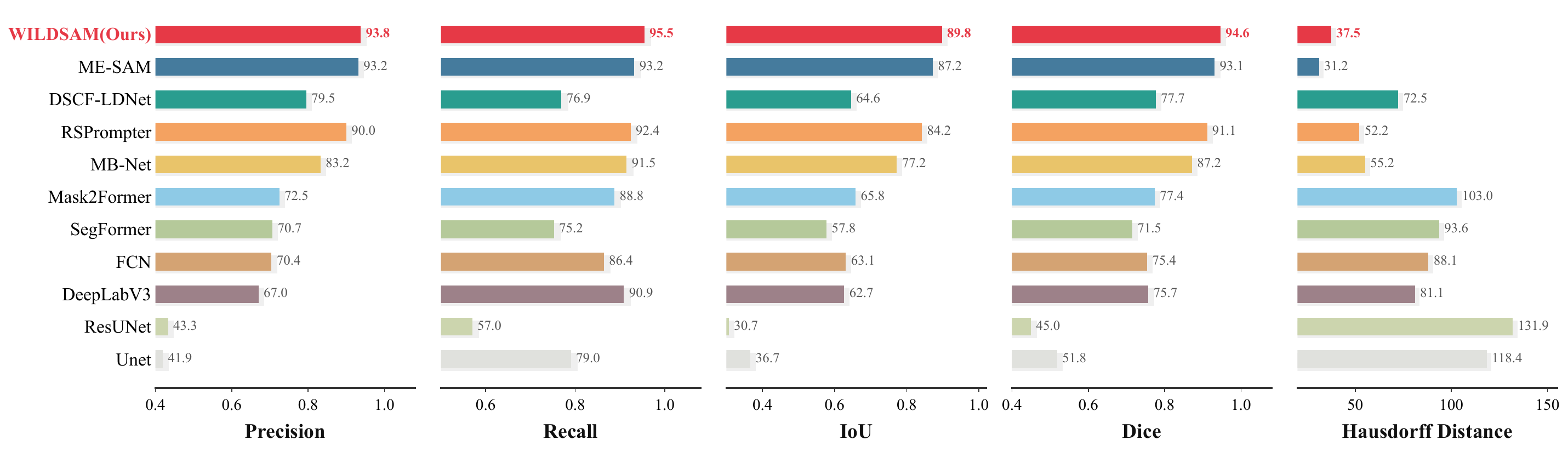}
\caption{Quantitative comparison of cross-region generalization performance on the Hunza-InSAR\cite{HUSSAIN2025103956} dataset across five evaluation metrics. For Precision, Recall, IoU, and Dice, higher values indicate better performance, whereas for the Hausdorff Distance (HD), a lower value denotes superior boundary accuracy. }
%\vspace{-.2cm}
    \label{fig:10}
\end{figure*}
% ====== figure ======
\begin{figure*}[!t]
    \centering
    \includegraphics[width=0.98\textwidth]{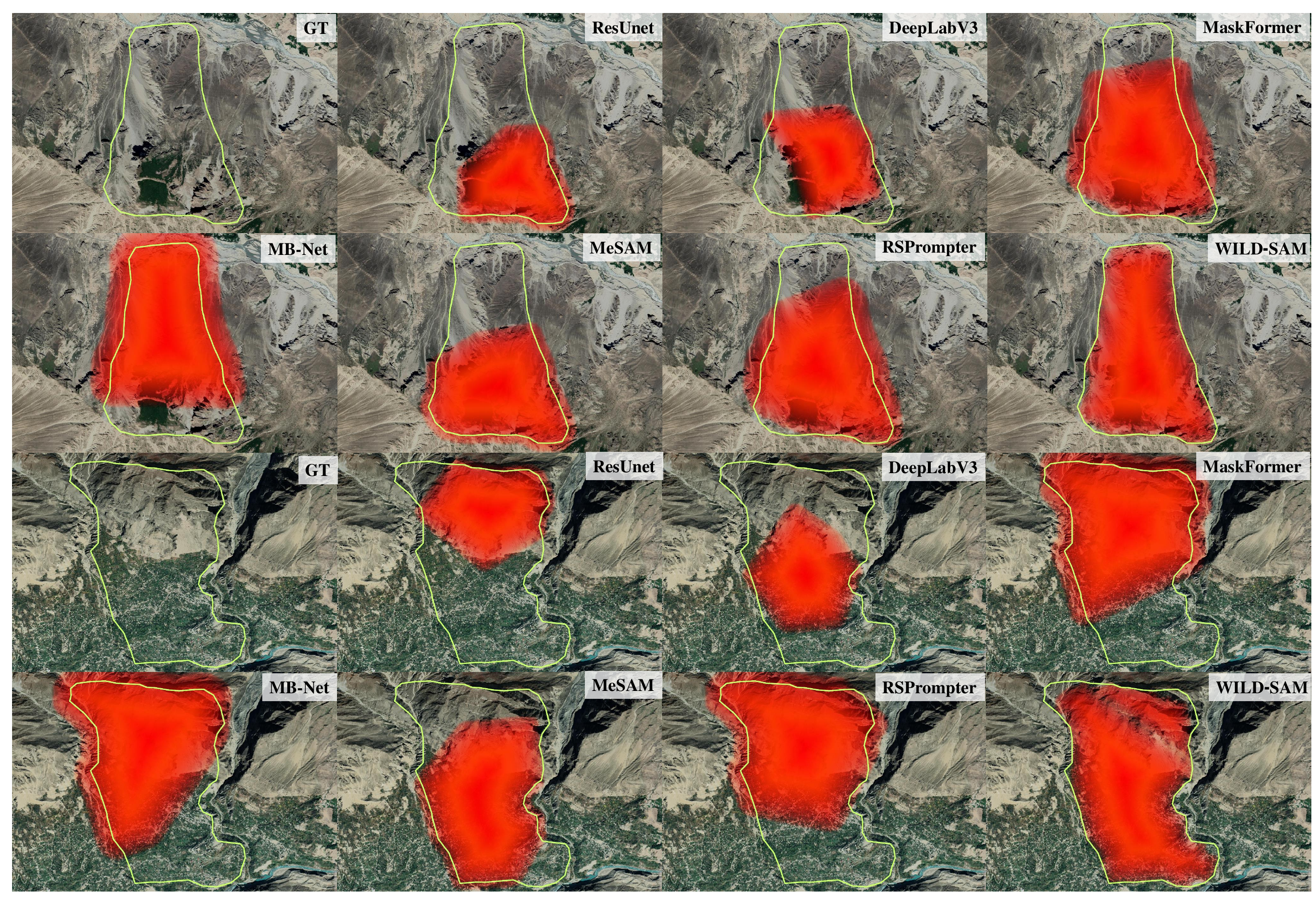}
    \caption{Cross-region qualitative evaluation on the external Hunza-InSAR\cite{HUSSAIN2025103956} dataset.}
    \label{fig:7}
\end{figure*}
\subsection{Visualization}  

To demonstrate the superior capability of WILD-SAM, we present a visual comparison against SOTA methods in Fig. \ref{fig:5}. Overall, WILD-SAM consistently yields segmentation masks that are topologically consistent with the ground truth. Specifically, for highly irregular targets with heavy speckle noise shown in the first row, competing methods like ResUnet and SAM-based RSPrompter exhibit fragmented segmentations. In contrast, WILD-SAM accurately reconstructs the continuous topology because the PA-MoE Adapter utilizes a dynamic routing mechanism to filter coherence noise while experts extract complex phase textures. Furthermore, for micro-targets with weak signals presented in the second row, generic networks and even specialized SOTA models suffer from severe false negatives. Conversely, WILD-SAM successfully captures these minute geometries as the Laplacian Edge Conv expert within the PA-MoE explicitly amplifies high-frequency phase shifts to circumvent the low-frequency bottleneck of the frozen backbone. When delineating closely spaced targets as observed in the third row, competitors such as DSCF-LDNet and MeSAM suffer from severe under-segmentation. WILD-SAM maintains clear instance separation because WGSE converts the coupled high-frequency features into dense prompts to sharpen decision boundaries. Finally, in low-contrast scenarios depicted in the fourth row, WILD-SAM achieves clean segmentations without the excessive false predictions seen in models like MB-Net. This robustness validates the PA-MoE, particularly the dilated convolution expert which expands the receptive field to ensure stable predictions regardless of signal-to-noise ratios.

To further validate the generalization capability of WILD-SAM across diverse and complex scenarios, we extend our visual comparison to the ISSLIDE+ dataset, as illustrated in Fig. \ref{fig:6}. For distinct diagonal and strip-like deformation features shown in the fourth row, SOTA architectures like MaskFormer and MeSAM fail to preserve directionality, often degrading the slender strip into a coarse blob. In strong contrast, WILD-SAM precisely reconstructs the smooth diagonal trajectory by leveraging the asymmetric convolution expert within the PA-MoE to model directional correlations. Moreover, regarding the identification of micro-targets embedded in low-contrast backgrounds in the second row, most competing networks suffer from feature vanishing and miss the target entirely. WILD-SAM accurately localizes these minute geometries with crisp boundaries through the synergy of the Laplacian expert and the WGSE strategy. When delineating dense clusters of adjacent deformation areas in the third row, a common failure mode is under-segmentation where methods such as SegFormer and MB-Net aggressively merge instances. WILD-SAM successfully separates them by utilizing cross-band attention within the WGSE to enhance high-frequency gap energies, thereby sharpening decision boundaries. Finally, in noisy interferograms with strong background fringes observed in the first row, MaskFormer suffers from severe over-segmentation while DSCF-LDNet produces fragmented masks. WILD-SAM achieves an optimal trade-off by capturing the complete shape while filtering out background fringes. This noise robustness is driven by the dilated convolution expert and the dynamic routing network, which adaptively suppress high-frequency experts in background regions to avoid noise amplification.

\subsection{Generalization Experiment}

To assess cross-region transferability under pronounced domain shifts, we evaluate all methods on the Hunza-InSAR dataset as an external generalization benchmark. As clearly illustrated in the multi-panel horizontal bar charts of Fig. \ref{fig:10}, WILD-SAM demonstrates superior generalization capabilities and achieves the most balanced performance across all criteria. Specifically, our method delivers an exceptional Precision of 0.9377 and an IoU of 0.8977. In stark contrast, while competitive baselines like MaskFormer maintain a relatively high Recall, they suffer a severe precision degradation to 0.7246 due to massive false alarms in the new geological domain. Even among the specialized models, WILD-SAM maintains a clear advantage. For instance, it significantly outperforms the landslide-specific MB-Net and DSCF-LDNet, which achieve lower IoU scores of 0.7725 and 0.6456 respectively. Although the SAM-based ME-SAM achieves the lowest HD of 31.18, WILD-SAM remains highly competitive with an HD of 37.51, substantially improving upon generic models like DeepLabV3 and MaskFormer that exhibit significantly elevated errors reaching 81.05 and 102.96 respectively. This robust cross-domain stability is fundamentally attributed to the PA-MoE Adapter. By dynamically routing multi-scale features, this module adaptively filters out coherence noise and varying artifacts in the new observation conditions, effectively preventing the frozen backbone from overfitting to the specific spectral distribution of the source domain.
% ====== figure ======
\begin{figure*}[t]
    \centering
    \includegraphics[width=\textwidth]{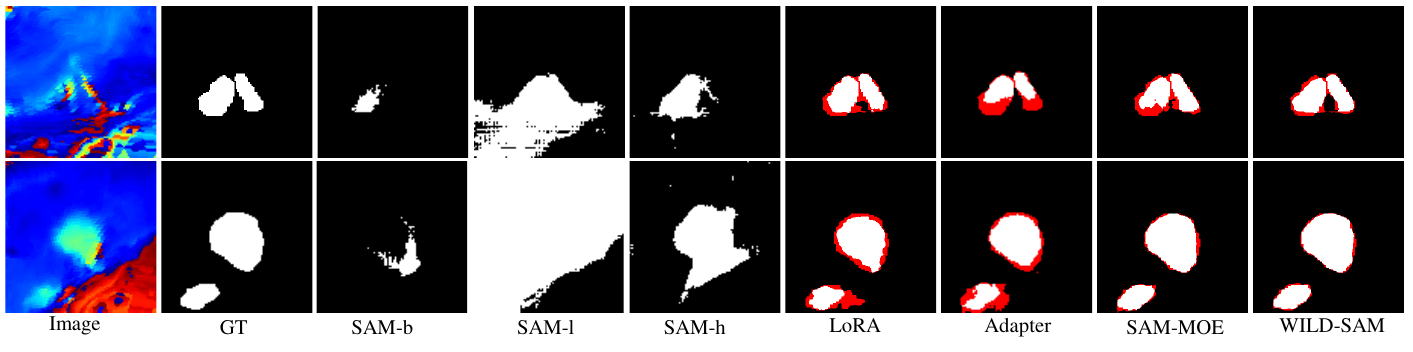}
    \caption{Qualitative ablation results on the ISSLIDE\cite{10433141} dataset. For the PEFT-based methods, the red regions highlight the discrepancy between the predicted masks and the ground truth.}
    \label{fig:8}
\end{figure*}
Qualitative visual comparisons depicted in Fig. \ref{fig:7} further validate that WILD-SAM maintains exceptional geometric consistency and contour fidelity when transferred to a completely unseen geological environment. For instance, when delineating large and highly irregular landslide bodies, baseline methods like MaskFormer and RSPrompter exhibit severe false expansions into surrounding stable slopes, resulting in heavily distorted and jagged boundaries. Conversely, WILD-SAM accurately localizes the active deformation areas with crisp and smooth contours that strictly align with the reference landslide outlines. This visual superiority confirms the efficacy of the WGSE strategy. By explicitly disentangling and injecting high-frequency wavelet subbands as dense prompts, WGSE compels the mask decoder to strictly adhere to physical phase discontinuities and sharp structural changes, effectively overcoming the unreliability of global semantic context in unfamiliar domains.
\subsection{Ablation Studies} 

To verify the contribution of each key design and the rationale behind its internal structures, we conduct extensive ablation studies on the ISSLIDE dataset. The quantitative results and qualitative visual comparisons are presented in Table \ref{tab:2} and Fig. \ref{fig:8}, respectively. To establish a rigorous baseline progression, we compare our full WILD-SAM model against zero-shot SAM variants, such as SAM-b, SAM-l, SAM-h and standard PEFT methods, such as LoRA and general Adapters.
\begin{table}[t]
\centering
\caption{Ablation study on the ISSLIDE\cite{10433141} dataset. $SAM$-$b$, $SAM$-$l$, and $SAM$-$h$ denote the SAM with Base, Large, and Huge ViT backbones, respectively. SAM-MOE represents the proposed WILD-SAM architecture without the WGSE module. The best values are shown in \textbf{bold}, and the second-best values are \underline{underlined}.}
\label{tab:2}
\renewcommand{\arraystretch}{1.1}
\setlength{\tabcolsep}{4.2pt}
\small
\resizebox{\columnwidth}{!}{%
\begin{tabular}{c | ccccc}
\noalign{\hrule height 1.5pt}
\textbf{Method} & \textbf{Prec (\%)} & \textbf{Rec (\%)} & \textbf{IoU ({\scriptsize $\uparrow$})} & \textbf{Dice ({\scriptsize $\uparrow$})} & \textbf{HD ({\scriptsize $\downarrow$})} \\
\hline
SAM-b          & 77.12          & 47.68          & 0.3602          & 0.4830           & 28.61 \\
SAM-l          & 65.71          & 61.48          & 0.3823          & 0.5046           & 32.42 \\
SAM-h          & 68.45          & 59.95          & 0.3915          & 0.5152           & 30.15 \\
LoRA\cite{10637992}           & 78.90          & 83.18          & 0.5368          & 0.6803           & 21.08 \\
Adapters\cite{chen2023sam}       & \underline{81.92} & 70.17          & 0.5980          & 0.7283           & 21.77 \\
SAM-MoE        & 81.15          & \underline{87.07} & \underline{0.7205} & \underline{0.8297}  & \underline{17.54} \\
\textbf{Ours}  & \textbf{83.15} & \textbf{88.19} & \textbf{0.7489} & \textbf{0.8484}  & \textbf{15.33} \\
\noalign{\hrule height 1.5pt}
\end{tabular}%
}
\end{table}
Under the severe domain shift of wrapped interferograms, the complete WILD-SAM framework achieves the best overall quantitative performance, successfully overcoming the limitations of zero-shot SAM variants and standard fine-tuning methods. Specifically, compared to the zero-shot SAM-h model which yields a low IoU of 0.3915 and a large JHD of 30.15, and the standard Adapters which only reach an IoU of 0.5980, the proposed WILD-SAM significantly boosts the region overlap to a Dice score of 0.8484 and substantially reduces the boundary error to an HD of 15.33. This significant enhancement is fundamentally attributed to the PA-MoE architecture, which enhances feature extraction under spectral and texture mismatch via expert routing, while the WGSE explicitly emphasizes phase discontinuities and sharp boundaries through wavelet-based high-frequency prompts.

Qualitative visual comparisons across representative cases in Fig. \ref{fig:8} demonstrate that WILD-SAM consistently generates segmentation masks with superior geometric continuity and contour fidelity compared to other baseline configurations. For instance, zero-shot SAM variants suffer from insufficient spatial coverage accompanied by contour drift, and standard LoRA or Adapters still produce jagged boundaries and scattered artifacts in heavily textured regions, whereas the full WILD-SAM model successfully suppresses isolated noise and yields sharper boundaries. This visual superiority is fundamentally attributed to the PA-MoE Adapter selectively aggregating discriminative spectral-texture cues via expert routing to create coherent masks, combined with the WGSE injecting fine-grained edge information into the decoding process to tightly constrain the target geometry.

\subsubsection{Effectiveness of the PA-MoE Adapter}  To assess the specific contribution of each heterogeneous expert within the CORE module, we conduct a progressive ablation study on the ISSLIDE and ISSLIDE+ datasets, as detailed in Table \ref{tab:3}.
The experimental results indicate that the representation capability of the model scales positively with the number of active experts, where configurations with three experts consistently outperform those with only two. Starting with the baseline combinations of two experts, such as $E1+E2$ or $E2+E3$, the model captures fundamental phase textures but exhibits limited precision in complex deformation areas. For instance, the $E2+E3$ configuration yields the lowest IoU of 0.7072 on ISSLIDE, primarily because it lacks both the foundational spectral prior from $E1$ and the high-frequency guidance from $E4$.
\begin{table}[t]
\centering
\caption{Ablation study of the PA-MoE adapter on the ISSLIDE\cite{10433141} and ISSLIDE+\cite{10433141} datasets. A checkmark (\checkmark) indicates the inclusion of the corresponding expert in the configuration: $E1$ (Depthwise), $E2$ (Dilated), $E3$ (Asymmetric), and $E4$ (Laplacian). The best values are highlighted in \textbf{bold}.}
\label{tab:3}
\renewcommand{\arraystretch}{1.2} 
\setlength{\tabcolsep}{3pt} 
\begin{tabular}{cccc|ccc|ccc}
\noalign{\hrule height 1.5pt} 
\multicolumn{4}{c|}{\textbf{CORE Settings}} & 
\multicolumn{3}{c|}{\textbf{ISSLIDE}} &       
\multicolumn{3}{c}{\textbf{ISSLIDE+}} \\
\hline
$E1$ & $E2$ & $E3$ & $E4$ & 
IoU ({\scriptsize $\uparrow$}) & Dice ({\scriptsize $\uparrow$}) & HD ({\scriptsize $\downarrow$}) & 
IoU ({\scriptsize $\uparrow$}) & Dice ({\scriptsize $\uparrow$}) & HD ({\scriptsize $\downarrow$}) \\
\hline

\checkmark & \checkmark & & & 
0.7115 & 0.8205 & 17.85 & 
0.8730 & 0.9195 & 9.86 \\

& & \checkmark& \checkmark& 
0.7085 & 0.8180 & 18.24 & 
0.8760 & 0.9205 & 8.45 \\

\checkmark & & & \checkmark & 
0.7128 & 0.8224 & 17.72 & 
0.8745 & 0.9208 & 9.15 \\

& \checkmark & \checkmark & & 
0.7072 & 0.8165 & 18.42 & 
0.8718 & 0.9175 & 10.12 \\

\hline

\checkmark & \checkmark & \checkmark & & 
0.7135 & 0.8215 & 17.52 & 
0.8775 & 0.9230 & 8.12 \\

& \checkmark & \checkmark & \checkmark & 
0.7282 & 0.8385 & 16.15 & 
0.8935 & 0.9382 & 7.08 \\
\hline

\checkmark & \checkmark & \checkmark & \checkmark & 
\textbf{0.7378} & \textbf{0.8467} & \textbf{15.35} & 
\textbf{0.9023} & \textbf{0.9469} & \textbf{6.28} \\
\noalign{\hrule height 1.5pt} 
\end{tabular}
\end{table}
The most significant performance increments are observed when the $E4$ expert, dedicated to Laplacian edge convolution, is integrated into the routing mechanism. Regardless of the specific combination, the inclusion of $E4$ leads to a mandatory improvement in segmentation precision. This is exemplified by comparing $E1+E2+E3$ and $E2+E3+E4$; despite both using three experts, the latter achieves a substantially higher IoU of 0.7282 on ISSLIDE and 0.8935 on ISSLIDE+. This quantitative breakthrough confirms that high-frequency phase discontinuities captured by the Laplacian operator are critical for delineating sharp landslide boundaries, effectively compensating for the low-frequency bias of the standard ViT backbone.

The full WILD-SAM framework, which orchestrates all four experts ($E1 \sim E4$) via the dynamic routing mechanism, achieves the optimal performance across all benchmarks. The synergy of heterogeneous experts allows the model to adaptively balance global context, directional semantics, and local boundary details. By dynamically suppressing coherence noise while simultaneously amplifying subtle deformation signals, the PA-MoE ensures that the frozen backbone accurately perceives the unique textural properties of interferometric phases, resulting in the best overall target completeness and contour fidelity.

\subsubsection{Effectiveness of WGSE}  
\begin{figure}[t]
    \centering
    \includegraphics[width=1\linewidth]{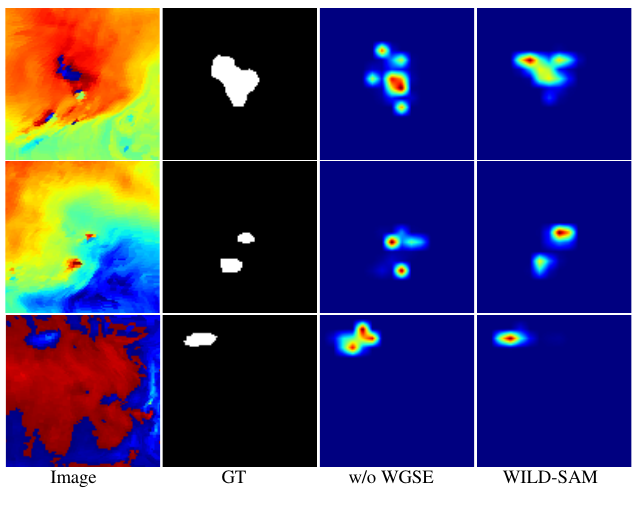}
\caption{Visualization of feature attention maps input to the Mask Decoder. From left to right: input image, ground truth (GT), attention map without WGSE, and attention map of the proposed WILD-SAM.}
%\vspace{-.2cm}
\label{fig:9}
\end{figure}

To intuitively verify the contribution of the WGSE strategy in recovering spatial details, we visualize the feature attention maps fed into the Mask Decoder in Fig. \ref{fig:9}.
In the baseline model, as shown in the third column, the attention focus is predominantly concentrated on the coarse semantic center of the landslide, while the boundary regions remain dimly activated. This phenomenon aligns with the "low-frequency bias" of the ViT backbone, which tends to smooth out high-frequency phase discontinuities, leading to under-segmentation or blurred edges.
In contrast, after incorporating the WGSE strategy , the model successfully activates the high-frequency cues along the landslide boundaries. By explicitly injecting wavelet-decomposed subbands as dense prompts, WGSE acts as a spectral sharpener, effectively guiding the decoder to attend to the subtle phase fringes and sharp edges that were previously suppressed.
This visual evidence corroborates that WGSE not only enriches the feature representation but also enforces a strong topological constraint, ensuring precise delineation of landslide geometries.

\begin{table}[t]
\centering
\caption{Comparison of model complexity, computational efficiency, and performance. For the proposed framework, the suffixes -S (Small), -B (Base), and -L (Large) denote configurations where the PA-MoE adapter is integrated into 3, 6, and 12 ViT encoder layers, respectively. Trainable parameters are reported in parentheses.}
\label{tab:4}
\renewcommand{\arraystretch}{1.2}
\setlength{\tabcolsep}{10pt}
\small
\begin{tabular}{c|ccc}
\noalign{\hrule height 1.5pt}
\textbf{Method} & \textbf{Params (M)} & \textbf{FLOPs (G)} & \textbf{Dice} \\
\hline
FCN\cite{long2015fully} & 32.95 & 34.72 & 0.5232 \\
DeepLabV3\cite{chen2017rethinking}  & 39.63 & 41.03 & 0.5283 \\
SegFormer\cite{xie2021segformer} & 44.70 & 54.70 & 0.6565 \\
MaskFormer\cite{cheng2021per} & 44.00 & 55.00 & 0.6670 \\
MB-Net\cite{ZHANG2024104300} & 72.24 & 67.91 & 0.7306 \\
RSPrompter\cite{10409216}  & 98.50 & 100.11 & 0.7103 \\
MeSAM\cite{rs16071150} & 93.00 & 72.21 & 0.7194 \\
\hline
WILD-SAM-S & 106.06 (19.39) & 68.11 & 0.7816 \\
WILD-SAM-B & 117.20 (30.53) & 103.80 & 0.8185 \\
\textbf{WILD-SAM-L} & \textbf{139.48} (\textbf{52.81}) & \textbf{175.18} & \textbf{0.8567} \\
\noalign{\hrule height 1.5pt}
\end{tabular}
\end{table}
\subsubsection{Model Complexity and Efficiency}  To evaluate the trade-off between model efficiency and segmentation performance, Table \ref{tab:4}. compares the total parameters, trainable parameters, floating-point operations (FLOPs), and Dice scores among different architectures. The proposed WILD-SAM framework achieves an optimal balance between computational cost and accuracy by leveraging a parameter-efficient fine-tuning strategy. Specifically, scaling the integration of the PA-MoE adapter from the lightweight WILD-SAM-S to the base WILD-SAM-B and full-scale WILD-SAM-L yields a progressive enhancement in segmentation accuracy, with Dice scores rising from 0.7816 to 0.8185 and peaking at 0.8567. Despite this consistent performance gain, the training overhead remains strictly constrained. Concurrently, the most compact WILD-SAM-S utilizes a mere 19.39 M trainable parameters to significantly outperform generic baselines such as SegFormer and MaskFormer. This scalable efficiency is fundamentally driven by the PA-MoE adapter paradigm, which freezes the massive ViT foundation backbone and selectively updates lightweight heterogeneous experts, effectively mitigating overfitting risks in data-constrained scenarios while ensuring precise interferometric phase interpretation.

\section{Conclusion}\label{sec:con} 

In this paper, we introduced WILD-SAM, a novel parameter-efficient fine-tuning framework that successfully adapts the SAM for high-precision wrapped interferometric landslide detection. Our architecture effectively bridges the profound spectral domain gap and overcomes the low-frequency bias of ViTs by integrating a PA-MoE Adapter and a WGSE strategy. Quantitative evaluations on the ISSLIDE benchmark demonstrate that WILD-SAM achieves SOTA target completeness and contour fidelity, boosting the Dice score to 0.8526 and reducing the HD to 15.33. This exceptional superiority is fundamentally attributed to the PA-MoE Adapter, which utilizes a dynamic routing mechanism to adaptively filter out variable coherence noise and maintain robust cross-domain generalization. Concurrently, the WGSE strategy explicitly leverages frequency-aware prompt injection to provide strict topological constraints, ensuring that the model delineates geometrically continuous boundaries even under severe phase noise. Ultimately, by bypassing computationally prohibitive phase-unwrapping processes, WILD-SAM enables reliable, automated, and large-scale geohazard monitoring directly from wrapped interferograms.

Despite the promising results, the current framework primarily operates on spatial-spectral features within individual interferograms. This reliance may limit its ability to effectively distinguish subtle deformation signals from atmospheric phase screens or turbulent noise, which often requires long-term temporal analysis to resolve.In future work, we plan to extend the WILD-SAM framework to time-series InSAR analysis, exploring temporal attention mechanisms to enhance consistency and robustness against atmospheric artifacts.

\vfill

\end{document}